\documentclass{article}

\usepackage{graphicx}
\usepackage{amsmath,amssymb} 
\usepackage{amsfonts}
\usepackage[dvipsnames]{color}
\usepackage{booktabs}
\usepackage{subfig}
\usepackage{xcolor,colortbl}

\usepackage[final]{style/corl_2020}

\usepackage{macros/mysymbols}
\usepackage{url}
\newcommand{\xhdr}[1]{\vspace{2pt}\noindent\textbf{#1}}

\usepackage{letltxmacro}

\LetLtxMacro{\oldsection}{\section}
\renewcommand{\section}[1]{
    \oldsection{#1}
    \vspace{-0.10in}
}

\LetLtxMacro{\oldsubsection}{\subsection}
\renewcommand{\subsection}[1]{
    \oldsubsection{#1}
    \vspace{-0.08in}
}

\LetLtxMacro{\oldsubsubsection}{\subsubsection}
\renewcommand{\subsubsection}[1]{
    \oldsubsubsection{#1}
    \vspace{-0.05in}
}

\title{Integrating Egocentric Localization for More Realistic Point-Goal Navigation Agents}

\author{
    \text{Samyak Datta}\textsuperscript{1}
    \And
    Oleksandr Maksymets\textsuperscript{2}
    \And
    Judy Hoffman\textsuperscript{1}
    \AND
    Stefan Lee\textsuperscript{2,3}
    \And
    Dhruv Batra\textsuperscript{1,2}
    \And
    Devi Parikh\textsuperscript{1,2}
    \AND
    \\
    \textsuperscript{1} Georgia Tech
    \And
    \\
    \textsuperscript{2} Facebook AI Research
    \And
    \\
    \textsuperscript{3} Oregon State University 
}

\begin{document}
\maketitle

\begin{abstract}
Recent work has presented embodied agents that can navigate to point-goal targets in novel indoor environments with near-perfect accuracy. However, these agents are equipped with idealized sensors for localization and take deterministic actions. This setting is practically sterile by comparison to the dirty reality of noisy sensors and actuations in the real world -- wheels can slip, motion sensors have error, actuations can rebound. In this work, we take a step towards this noisy reality, developing point-goal navigation agents that rely on visual estimates of egomotion under noisy action dynamics. We find these agents outperform naive adaptions of current point-goal agents to this setting as well as those incorporating classic localization baselines. Further, our model conceptually divides learning agent dynamics or odometry (where am I?) from task-specific navigation policy (where do I want to go?). This enables a seamless adaption to changing dynamics (a different robot or floor type) by simply re-calibrating the visual odometry model -- circumventing the expense of re-training of the navigation policy. Our agent was the runner-up in the PointNav track of CVPR 2020 Habitat Challenge.

\end{abstract}
\section{Introduction}

Impressive progress has been made in training agents to navigate inside photo-realistic, 3D simulated environments \cite{habitat19iccv}. One of the most fundamental and widely studied of these tasks is Point-Goal navigation (PointNav) \cite{anderson2018evaluation}. In PointNav, an agent is spawned in a never-before-seen environment and asked to navigate to target coordinates specified relative to the agent's start position. No map is provided. The agent must navigate from egocentric perception and stop within a fixed distance of the target to be successful. Recent work \cite{ddppo} has developed agents that can perform this task with near-perfect accuracy -- not only achieving success 99.6\% of the time but also following \emph{near-shortest paths} from start to goal while doing so, all in novel environments without a map! 

\begin{figure}[t]
    \centering
        \includegraphics[scale=0.17]{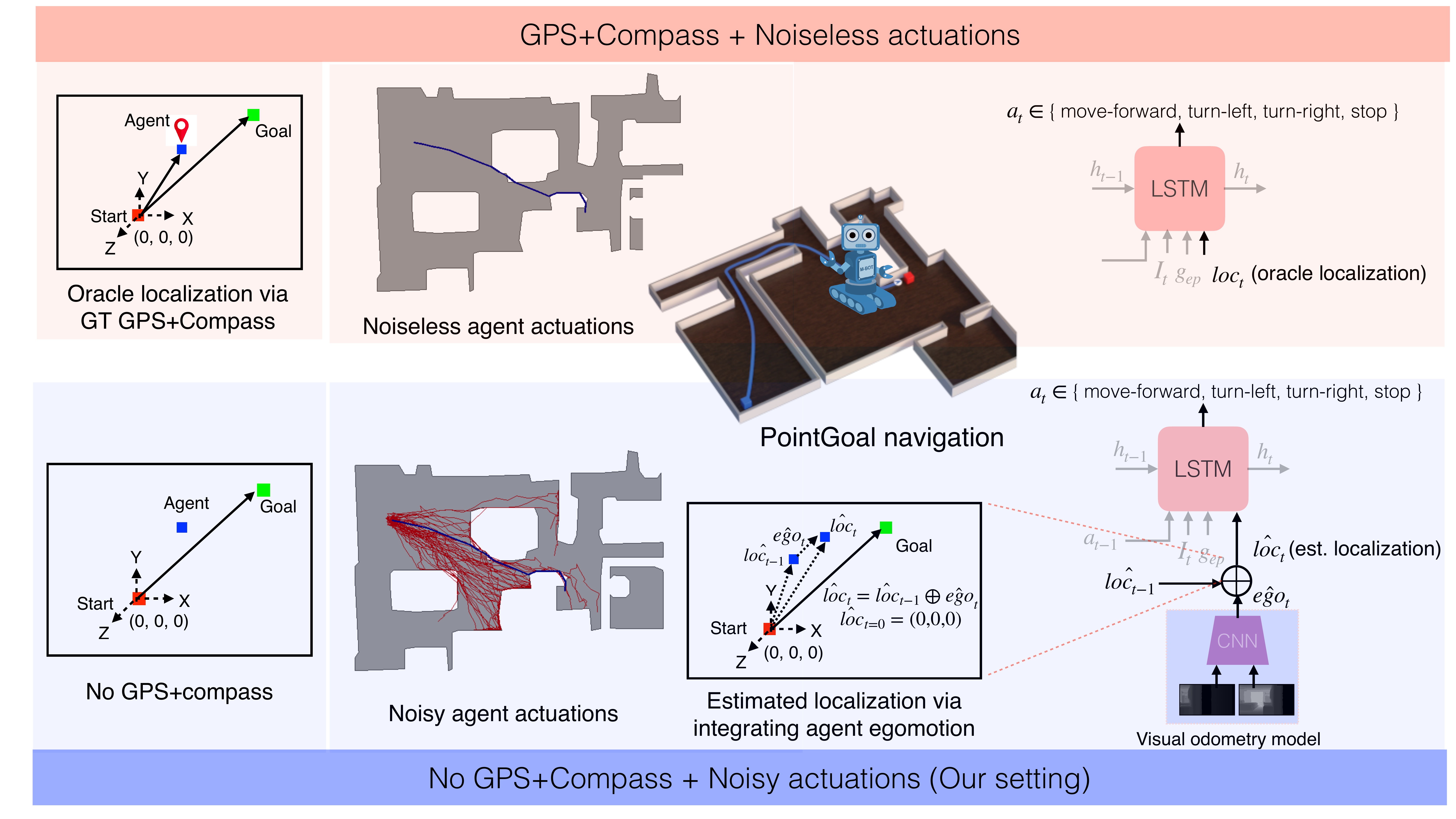}
    \caption{We lift assumptions of access to perfect localization via an idealized GPS+Compass sensor and noise-free agent actuations for Point-Goal navigation. Our proposed agent integrates learned visual odometry predictions to derive an estimate of its location in the trajectory. We train and evaluate our agent in environments with both noise-free and noisy actuations.}
    \label{fig:teaser}
\end{figure}

While commendable, this result is not without some caveats. The standard PointNav setting (as in \cite{ddppo, habitat_challenge}) provides agents with perfect localization, which is referred to as an idealized `GPS+Compass' sensor in that literature. Moreover, all agent actions are executed without any error or noise. Unfortunately, having accurate and precise localization information in unseen and previously unmapped indoor environments is unrealistic given current technology.
Likewise, agent actuation is inherently noisy due to constraints such as imperfections in physical parts,
variations in the environment, 
(\eg a worn spot on the carpet that provides less friction),
and the chaotic result of collisions with objects. 

In addition to being unrealistic, these assumptions also shortcut the hard problems that a real robot must learn to solve this task. For example, oracle localization provided by the environment enables agents to continuously update the relative coordinates to the goal location -- trivializing the problem of identifying when the goal has been reached.

In this work, we lift these assumptions and explore the problem of PointNav under noisy actuation and without oracle localization (no `GPS+Compass' sensor). We develop models with internal visual odometry modules that estimate egomotion (changes in position and heading) from consecutive visual observations. This allows the agent to maintain a noisy, but up-to-date estimate of pose by integrating the per-action egomotion estimates along its trajectory (Fig. \ref{fig:teaser}). Overall, this agent is provided the point-goal location in relative coordinates \emph{once} at the start of each episode and must rely entirely on its own estimations to navigate effectively. We analyze our agent in two settings: without and with actuation noise. The former matches experimental protocols of prior work \cite{habitat19iccv,ddppo}; the latter is our contribution and more realistic.

In both cases, we show that our agent \emph{significantly} outperforms an entire range of highly competitive baselines. In particular, we compare against a ``dead-reckoning'' baseline that relies on a fixed mapping from actions to odometry readings to estimate pose. In the absence of actuation noise, our approach involving \textit{learning} as opposed to \textit{memorization} of odometry estimates outperforms dead-reckoning with a relative improvement of 68\% in SPL ($0.51$ vs. $0.30$). Note that dead-reckoning serves as a very strong baseline for noiseless actuation: an agent equipped with knowledge about the environment dynamics  
can maintain an accurate estimate of its location in a seemingly straightforward manner. However, our empirical results show that dead-reckoning estimates are poor due to collisions, leaving the agent with no avenues to recover from even a single faulty localization. We also compare to and outperform adaptions of current PointNav agents that are re-trained to operate without localization information. On account of being monolithic neural policies trained end-to-end for the task, these adaptions, in principle, could learn to implicitly integrate egomotion. However, our approach, that involves an explicit module for predicting and integrating egomotion, outperforms these adaptions with relative improvements of over $466\%$ (SPL=$0.51$ v/s SPL=$0.09$).

In the noisy setting, we again outperform dead-reckoning and adaptions of current PointNav agents with relative improvements of 194\% and 114\%, respectively. In addition to that, we show that having a conceptual separation between the learning of agent dynamics (via the odometry module) and task-specific navigation policy allows for a seamless transfer of the latter from noiseless to noisy actuation environments. We show this via a simple fine-tuning of the learned odometry model (while keeping the underlying policy frozen) to adapt to the noisy actuations in the new setting. This is akin to a `hot-swappable' navigation policy that can be incorporated with different egomotion models, thereby circumventing the expensive re-training of the former. This is highly significant because prior work \cite{ddppo} requires billions of frames of experience, hundreds of GPUs, and over 6 months of GPU-time to train a near-perfect navigation policy. Our `hot-swapping' approach allows us to seamlessly leverage that work in noisy actuation settings with zero (re-)training of the policy. 

To summarize our contributions, to the best of our knowledge, we develop the first approach for PointNav in realistic conditions of noisy agent actuations and no localization (GPS+Compass sensor) input. A straightforward modification of our proposed agent was the runner-up for the PointNav track of the CVPR 2020 Habitat Challenge. We view this as a step towards one of the grand goals of the community -- making navigation agents more suitable for deployment  in the real world.
\section{Related Work}
\noindent\textbf{Goal-Driven Navigation.} The topic of goal-driven embodied navigation has witnessed exciting progress \cite{habitat19iccv,ddppo,das2018embodied,das2018neural,wijmans2019embodied,anderson2018vision}. Point-Goal navigation \cite{anderson2018evaluation}, being one of the most fundamental among these task, has also been the subject of several prior works. Point-Goal navigation approaches can be broadly categorized into two parts. \cite{gupta2017cognitive,chaplotmodular,parisotto2017neural} take inspiration from a classical decomposition of the navigation problem into building map representations and path-planning. More recently, \cite{habitat19iccv,ddppo} study end-to-end training of neural policies using RL. Among the above, \cite{gupta2017cognitive,parisotto2017neural,habitat19iccv,ddppo} either assume known agent egomotion or perfect localization. Furthermore, \cite{chaplotmodular} assumes access to a noisy variant of the agent’s pose obtained by manually adding noise (sampled from real-world experiments on LoCoBot \cite{murali2019pyrobot}) to the oracle sensor. Similarly, \cite{chen2018learning} uses a noisy variant of dead-reckoning to estimate agent location while building allocentric spatial maps for exploration. In contrast to these works, our agents operate in a more realistic setting which doesn’t assume any form of localization information whatsoever and has noisy transition dynamics. Furthermore, we show that our approach of integrating visual odometry predictions to estimate localization outperforms dead-reckoning baselines.

The motivations of our work are also closely related to \cite{watkins2019learning} where the authors train an agent to navigate to $8$-image panoramic targets without an explicit map or compass and using imitation learning on expert trajectories. In contrast, we work with a different task wherein our agents navigate to 3D goal locations as opposed to a panoramic image of the target in an unseen environment. Likewise, the authors in \cite{chen2019audio} show that binaural audio signals in indoor environments can serve as an alternate form of localization in the absence of GPS+Compass. In contrast, our method doesn't rely on the introduction of any additional modalities beyond vision to estimate localization.

\xhdr{Localization and Visual Odometry.} Our work is also related to the large body of work \cite{weyand2016planet,kendall2015posenet,wang2017deepvo} on localization and visual odometry. 
Most similar to our approach for training our agent's odometer is \cite{wang2017deepvo} that estimates relative camera pose from image sequences. However, unlike these models that are trained specifically for the end-task of odometry estimation, our goal is to integrate the visual odometry model into the specific downstream task of Point-Goal navigation. In this context, our egomotion estimator is used in an active, embodied set-up wherein the egocentric localization estimates are used by the agent to select actions that in turn affect the egomotion.

\xhdr{Reducing the Sim2Real Gap.} Our motivation to move towards more realistic PointNav settings is also aligned with \cite{habitatsim2real19arxiv}. The authors attempt to find the optimal simulation parameters such that improvements in simulation also translate to improvements in reality for Point-Goal navigation (with GPS and compass). In order to run in-reality experiments, they mounted a high-precision (and costly) LIDAR sensor to the robot in order to provide location information -- on-board IMUs, motor encoders, and wheel rotation counters proved too noisy. In contrast, our proposed approach performs navigation without the need for any expensive localization sensors.
\section{Approach}

\xhdr{Preliminaries.} We start by describing the Point-Goal navigation task with oracle localization and current agent architectures for the task, before describing the details of our approach.

\label{subsec:prelim}
In Point-Goal navigation \cite{anderson2018evaluation}, an agent is spawned at a random pose in an unseen environment and asked to navigate to goal coordinates specified relative to its start location. At every step of the episode, the agent receives RGB-D observation inputs via its visual sensors and a precise estimate of its location in the trajectory (relative to the start of the episode) via an idealized GPS+Compass sensor. Using these inputs for the current step, the agent performs an action by predicting a distribution over a discrete action space -- \{\texttt{move-forward}, \texttt{turn-left}, \texttt{turn-right}, \texttt{stop}\}.

Prior work \cite{habitat19iccv,wijmans2019embodied} has trained recurrent neural policies for Point-Goal navigation end-to-end using RL. At a high-level, these recurrent policies (modelled as a 2-layer LSTM) typically  predict actions based on the (ResNet-50) encoded representation of the current visual observation, its previous action, the goal coordinates for the episode and the noise-free localization estimate from the GPS+Compass sensor. Note that given the information about (a) the goal coordinates and (b) its current location both with respect to the start of the episode, it is straightforward for the agent to derive a noise-free estimate of the goal coordinates relative to its current state at every point in time.

\xhdr{Point-Goal Navigation without Oracle Localization.}
Our proposed approach removes the oracle localization information (GPS+Compass sensor) and instead equips agents with an odometry module that is responsible for predicting per-action egomotion estimates. Access to such an odometer allows the agent to integrate its egomotion estimates over the course of its trajectory -- thereby deriving a potentially erroneous substitute of the localization information.
Taking inspiration from existing literature in visual odometry, we train our odometry model to regress to the change in pose between two sequential observations. In the sections that follow, we describe the details of our visual odometry model, the dataset collection protocol that was followed to collect data for training the odometer and its integration with the agent policy.

\xhdr{Visual Odometry Model.}
\label{subsec:odometry-model}
We design our odometry model as a CNN that takes visual observation pairs (in our case, depth maps) as input and outputs an estimate of the relative pose change between the two states. We characterize the pose change as the positional offset ($\Delta x$, $\Delta y$, $\Delta z$) and change in heading/yaw ($\Delta \theta$) of the second state with respect to the first (Fig. \ref{fig:ego-dataset} (a) shows the agent coordinate system). More concretely, let $I_t$ and $I_{t+1}$ be the depth images corresponding to the agent states at time $t$ and $t+1$, respectively. Both frames are depth concatenated and passed through a series of 3 convolutional layers: $($Conv 8$\times$8, ReLU, Conv 4$\times$4, ReLU, Conv  3$\times$3, ReLU$)$. 
Subsequent fc-layers generate the flattened $512$-d embedding for the depth map pair, followed by predictions for $\Delta x$, $\Delta y$, $\Delta z$, and $\Delta \theta$ (egomotion deltas). The egomotion CNN is trained to regress to the ground-truth egomotion deltas by minimizing the smooth-L1 loss.

\xhdr{Egomotion Dataset.}
\label{subsec:dset}
\label{subsec:dataset}
To train our odometry model, we require a dataset comprising of observation pairs and the corresponding ground-truth egomotion between the two states defined by the observation pairs. These constitute the inputs and regression targets for the odometry CNN respectively. 

In order to collect this dataset, we adopt the follwing protocol. We take a Point-Goal navigation agent that has been trained with oracle localization and use its policy to unroll trajectories. Then, we sample pairs of source (src) and target (tgt) agent states from within those trajectories uniformly at random. For each (src, tgt) pair of states, we record (a) the corresponding visual observations ($v_t$ and $v_{t+1}$) and (b) the ground-truth 6-DoF camera pose from the simulator. The latter comprises of a translation vector, $t \in {\rm I\!R}^3$, denoting the agent's location in world coordinates and a rotation matrix, $R \in SO(3)$, representing a transformation from agent's current state to the world. Therefore, for a given pair of (src, tgt) states, we can obtain our egomotion transformation between (src, tgt) as: $T_{\text{tgt} \rightarrow \text{src}} = T_{\text{src} \rightarrow \text{world}}^{-1} \cdot T_{\text{tgt} \rightarrow \text{world}}$, where $T_{a \rightarrow b} = \begin{bmatrix}
R_{a \rightarrow b} & t_{a \rightarrow b} \\
0 & 1
\end{bmatrix}$.

In the absence of noise in agent actuations, we would, in principle, expect to obtain a deterministic mapping between agent actions and odometry readings. However, in practice, the agent occasionally suffers displacements along x while executing \texttt{move-forward} (see Fig. \ref{fig:ego-dataset}(a) for agent coordinate system). This happens due to collisions with the environment and in such cases, the agent's displacement along z i.e. its heading is also different than the standard setting of $0.25$m. \texttt{turn-left} and \texttt{turn-right} always correspond to $10^\circ$ (or $0.17$ radians).

We sample a total of ~1000 (src, tgt) pairs from each of the 72 scenes in the Gibson training split \cite{habitat19iccv} for a total of 72k pairs. Within each scene, we balance out the sampling of the ~1000 points by collecting an equal number of points from each of the 122 trajectories in that scene. Since our dataset has been collected by sampling from unrolled agent trajectories, the distribution of actions is representative of the type of actions an agent might take while navigating -- 58\%, 21\% and 21\% of the dataset corresponds to \texttt{move-forward}, \texttt{turn-left} and \texttt{turn-right} actions respectively.

\begin{figure}[t]
    \centering
        \includegraphics[scale=0.20, trim=0cm 1cm 0cm 0cm]{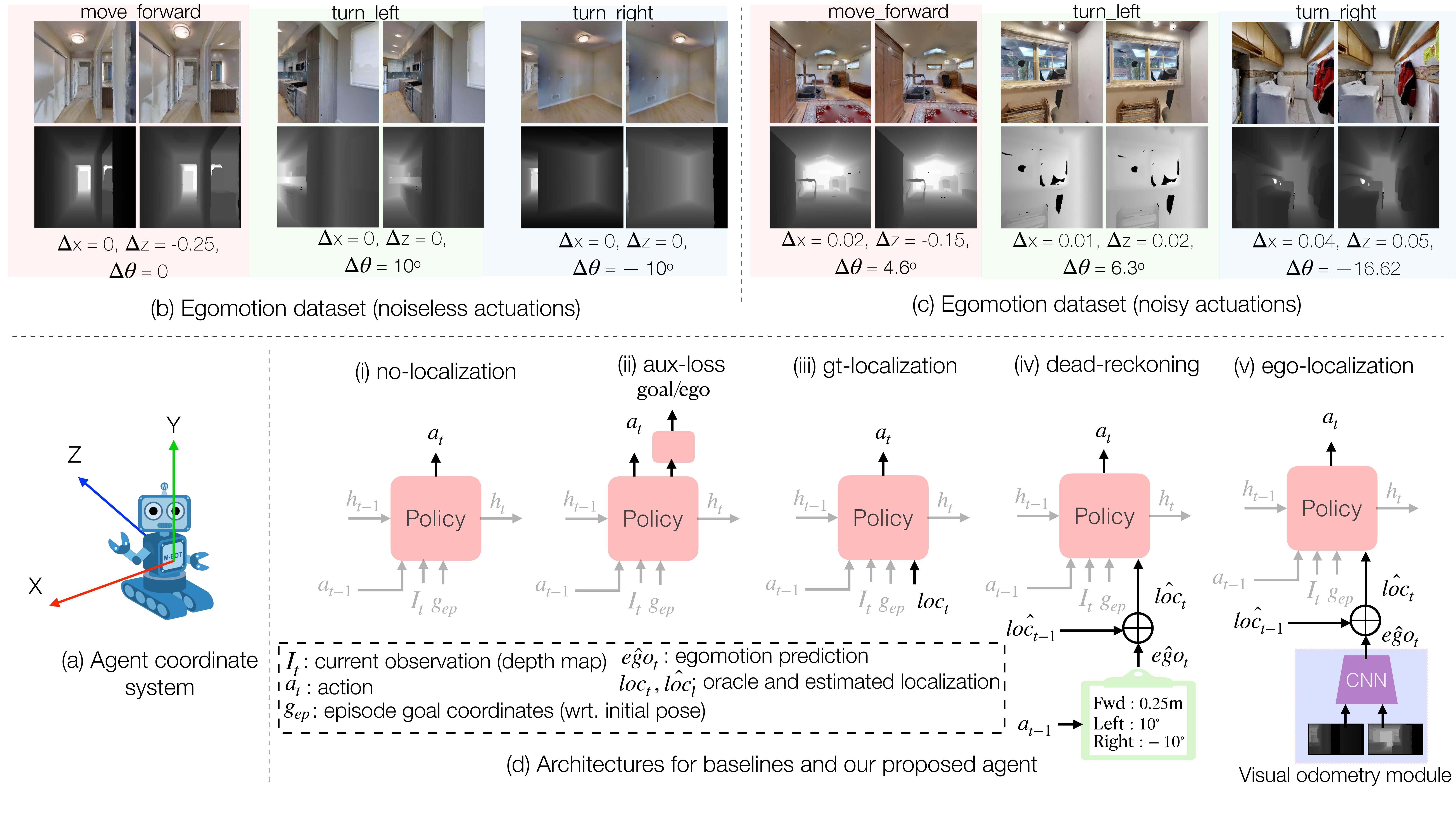}
    \caption{(a) shows the agent coordinate system. We also show samples of data points from the egomotion dataset collected under (b) noiseless and (c) noisy actuation settings. Each data point consists of visual observations (RGB-D maps) along with ground-truth egomotion ($\Delta \theta$ represents a rotation around Y) between two states defined by the observations. (d) We show architectures for our proposed (ego-localization) and baseline agents.}
    \label{fig:ego-dataset}
\end{figure}

\xhdr{Integrating Egomotion Estimation with the Agent.}
Having described the odometry model and the dataset that it's trained on, we now discuss how it integrates into the agent architecture. Recall (from Sec.\,\ref{subsec:prelim}) that, owing to the availability of perfect localization, the Point-Goal navigation agents from \cite{habitat19iccv,ddppo} can derive the relative goal coordinates at every state. In the absence of GPS+Compass, our proposed agent uses the odometry model to first estimate its location in the trajectory and subsequently, predicts the relative goal coordinates as follows.

At every step, the egomotion predictions from the odometry CNN are first converted to a $4\times4$ transformation matrix that represents a transformation of coordinates between the agent's previous and current states in the trajectory (using the equation above). The transformation is then applied to the relative goal coordinate predictions from the previous time step to project them on to the coordinate system defined by the agent's current state. This generates the relative goal coordinate input for the current step of the policy network.

\xhdr{Noisy actuations.}
\label{subsec: noisy-acts}
Real-world robot locomotion is far from deterministic due to the presence of actuation noise. In order to model such noisy agent actions, we leverage noise models derived from real-world benchmarking of an actual, physical robot by the authors in \cite{habitatsim2real19arxiv}. Specifically, we use both linear and rotational noise models from LoCoBot \cite{locobot}. The former is a bi-variate Gaussian (with a diagonal covariance matrix) that models egomotion errors along the z and x axis whereas the latter is a uni-variate Gaussian modeling egomotion errors in the agent's heading. For any given action, we perform truncated sampling and add noise from both the linear and rotational noise models.

We collect a version of the egomotion dataset under the noisy actuation setting by following the same protocol as described in Sec. \ref{subsec:dset}. Under this setting, the agent suffers non-zero egomotion deltas along all degrees of freedom for all action types. Please refer to the Supplementary document for details regarding the parameters of actuation noise models and a comparison of the distributions of per-action egomotion deltas in the dataset across noiseless and noisy actuation set-ups.

\xhdr{Training.}
\label{subsec:training}
The odometry CNN is pre-trained on the egomotion dataset and kept frozen. We train the agent's navigation policy using DD-PPO \cite{ddppo} -- a distributed, decentralized and synchronous implementation of the Proximal Policy Optimization (PPO) algorithm.

Let $d_t$ be the geodesic distance to target at time step $t$. Furthermore, let $s$ denote the terminal success reward obtained at the end of a successful episode (with $\mathbb{I}_{\text{success}}$ being the indicator variable denoting episode success) and $\lambda$ be a slack-reward to encourage efficient exploration. Then, the reward $r_t$ obtained by the agent at time $t$ is given by: $r_t = \underbrace{s\,.\,\mathbb{I_{\text{success}}}}_{\text{success reward}} + \underbrace{(d_{t-1} - d_t)}_{\text{reward shaping}} + \underbrace{\lambda}_{\text{slack reward}}$.
We set $s = 1.0, \lambda = -0.01$ for all our experiments. We train all our agents for a maximum of 60M frames. We initialize the weights of the visual encoder with those from an agent that has been trained (using ground-truth localization) for ~2.5B frames. Check Supplement for additional training details.
\section{Baselines}
\textbf{no-localization}: This is a naive adaption of the Point-Goal agents from \cite{habitat19iccv,ddppo} to our setting with no localization information. The policy network for the agents in \cite{habitat19iccv,ddppo} take the previous action, visual observations, episode goal and oracle GPS+Compass as inputs at every step (Sec. \ref{subsec:prelim}). For adapting this model to our setting, we drop the GPS+Compass input (keeping the other 3 unchanged). We train this agent with the same reward settings and losses, as described in Sec \ref{subsec:training} (and consistent with \cite{habitat19iccv,ddppo}). The performance of this baseline tells us how well an agent would do if it is trained for the Point-Goal navigation task using the state-of-the-art approach \cite{ddppo} but without any localization information whatsoever to guide its navigation.

\noindent \textbf{aux-loss (aux-loss-goal + aux-loss-ego)}: These agents are similar in architectural set-up to the no-localization baseline. However, they are trained with additional auxiliary losses that encourage them to predict information pertinent to their localization. At every step, the policy network is additionally trained to predict, from its hidden state, either the goal coordinates relative to its current state (\textbf{aux-loss-goal}) or the relative change of pose between its previous and current states (\textbf{aux-loss-ego}). Note that these baselines have access to the GPS+Compass sensor during training, just as our approach does to train our odometry module. But, they use this information indirectly as auxiliary losses rather than directly as localization information. Both our approach and these baselines do not use the oracle GPS+Compass information at test time. Comparing our approach to these baselines demonstrates the value of an explicit odometry module as opposed to relying on a monolithic neural policy agent to learn that it should infer localization-related information from auxiliary signals.

\noindent \textbf{dead-reckoning}: The agent derives localization estimates using a static look-up table that maps actions to associated odometry readings -- \texttt{move-forward}: displacement of $0.25$m along heading, \texttt{turn-left}/\texttt{turn-right}: $10^\circ$ on-spot rotation. A comparison with this baseline answers the question — is it really necessary to \emph{learn} the odometry estimates instead of naively memorizing them?

\noindent \textbf{classic}: We also compare with a classic robotic navigation pipeline that has modular components for map creation, localizing the agent on the map, planning a path, selecting a waypoint on the planned path and moving the agent along the predicted waypoint. Such pipelines have been extensively used in robotics with several choices available for each component. Following \cite{habitat19iccv}, we leverage prior work \cite{mishkin2019benchmarking} that proposes a complete implementation of such a pipeline that can be readily deployed in simulation. \cite{mishkin2019benchmarking} uses ORB-SLAM2 \cite{mur2017orb} as the agent localization module. We use the same set of hyperparameters as reported in the original work (refer to the Supplementary for more details).

\noindent \textbf{ego-localization}: Finally, this is our approach wherein the agent uses a trained odometry model to derive localization estimates for navigation using a neural policy.
    
As a benchmark for upper-bound performance, we also report numbers for the agent that navigates in the presence of `oracle' localization (\textbf{gt-localization}) during both training and test episodes.
\section{Results and Findings}

\xhdr{Environment.}
We use the Habitat simulator \cite{habitat19iccv} with the Gibson dataset of 3D scans \cite{gibson2018} for our experiments. We leverage the human-annotated ratings (on a scale of $1$-$5$) provided by \cite{habitat19iccv} for mesh reconstructions and only use high-quality environments ($\geq 4$ rating) for our experiments. For a given scene, we use the dataset of Point-Goal navigation episodes from \cite{habitat19iccv}. Overall, we work with $8784$ episodes across 72 scenes in train and 994 episodes across 14 scenes in val.
    
\xhdr{Metrics.}
\label{subsec:metrics}
A navigation episode is deemed successful if the agent calls \texttt{stop} within a distance of $0.2$m from the target location. We measure \textbf{success rate} as the fraction of successful episodes. Following prior work on Point-Goal navigation, we also report performance on the \textbf{Success Weighted by Path Length (SPL)} metric \cite{anderson2018evaluation}, averaged across all episodes in the validation split.

Owing to their reliance on a binary indicator of episode success, both success rate and SPL do not provide any information about episodes that fail. Consider the scenario (Fig \ref{fig:top-down-trajs}(c)) wherein an agent follows a path that closely resembles the shortest path between start and goal locations, but prematurely calls \texttt{stop} right at the 0.2m success perimeter boundary. Both success and SPL for the episode are $0$. Although the agent managed to navigate reasonably well, its performance gets harshly ignored from the overall statistics. Hence, relying on success and SPL as the sole metrics (as done by prior work) often leads to an incomplete picture of the agent’s overall performance. We observe that this problem is even more acute in set-ups without GPS+Compass where noisy pose estimates directly affects the agent's decision to call \texttt{stop} at the right distance relative to the goal.

To get a more holistic estimate of the agent's performance, we also report the geodesic distance to target upon episode termination (\textbf{geo\_d\_T}). In addition to that, we propose a new metric called \textbf{SoftSPL}. Let $d_{\text{init}}$ and $d_{\text{T}}$ denote the (geodesic) distances to target upon episode start and termination. The SoftSPL for an episode is defined as: $\text{SoftSPL} = \left( 1 - \frac{d_{\text{T}}}{d_{\text{init}}} \right) \cdot \left( \frac{s}{max(s, p)} \right)$
where $s$ and $p$ are the lengths of the shortest path and the path taken by the agent. SoftSPL replaces the binary success term in SPL with a ``soft'' value that is indicates the progress made by the agent towards the goal (``progress'' can be negative if the agent ends up farther away from the goal than where it started).

\definecolor{Gray}{gray}{0.90}
\newcolumntype{a}{>{\columncolor{Gray}}c}

\begin{table}[t]
    \centering
    \resizebox{1.0\linewidth}{!}{
        \parbox{0.8\linewidth}{
            \begin{tabular}{c a c c c c}
                & \multicolumn{5}{c}{Noiseless actuations}  \\
                \cmidrule{2-6}
                Agent & SoftSPL $\uparrow$ & SPL $\uparrow$ && Succ. $\uparrow$ &  geo\_d\_T $\downarrow$ \\
                \midrule
                no-localization & 0.726 & 0.096 && 0.099 & 1.573 \\
                aux-loss-goal & 0.758 & 0.179 && 0.192 & 1.117 \\
                aux-loss-ego & 0.680 & 0.086 && 0.090 & 1.825 \\
                dead-reckoning & 0.797 & 0.303 && 0.311 & 1.047 \\
                classic & 0.584 & 0.478 && \textbf{0.708} & 1.183 \\
                \cmidrule{1-6}
                ego-localization (Ours) & \textbf{0.813} & \textbf{0.508} && 0.535 & \textbf{0.959}\\
                \midrule
                GT-localization & 0.865 & 0.866 && 0.948 & 0.388 \\
                \bottomrule
            \end{tabular}
            \\ [19pt]
            \caption{We report Point-Goal navigation metrics for our proposed agent, baselines and ``oracle'' navigators for noiseless (above) and noisy (right) actuation settings.}
            \label{tab:results}
        }
        \quad
        \parbox{0.8\linewidth}{
            \begin{tabular}{c a c c c c}
                & \multicolumn{5}{c}{Noisy actuations}  \\
                \cmidrule{2-6}
                Agent & SoftSPL $\uparrow$ & SPL $\uparrow$ && Succ. $\uparrow$ & geo\_d\_T $\downarrow$ \\
                \cmidrule{1-6}
                
                \multicolumn{6}{c}{\textbf{no-localization}}  \\
                \multicolumn{1}{c}{old-policy} & 0.491 & 0.036 && 0.047 & 2.259 \\
                \multicolumn{1}{c}{re-trained-policy} & 0.515 & 0.022 && 0.028 & 2.318 \\
                \cmidrule{1-6}
                
                \multicolumn{6}{c}{\textbf{estimated-localization}}  \\
                 \multicolumn{1}{c}{old-policy} & 0.28 & 0.012 && 0.019 & 3.349 \\
                \multicolumn{1}{c}{re-trained-odom} & 0.494 & 0.044 && 0.058 & 2.000 \\
                \multicolumn{1}{c}{re-trained-odom+policy} & 0.576 & 0.047 && 0.060 & 1.843 \\
                 \multicolumn{1}{c}{dead-reckoning} & 0.407 & 0.016 && 0.023 & 2.767 \\
                 \multicolumn{1}{c}{classic} & 0.301 & 0.267 && 0.700 & 1.748 \\
                \midrule

                \multicolumn{6}{c}{\textbf{gt-localization}}  \\
                \multicolumn{1}{c}{old-policy} & 0.666 & 0.593 && 0.842 & 0.328 \\
                \multicolumn{1}{c}{re-trained-policy} & 0.671 & 0.650 && 0.924 & 0.266 \\
                \bottomrule
               
            \end{tabular}
        }
        
    }
    \label{tab:noiseless-act-results}
\end{table}

We now present navigation results for our proposed model and compare with baselines in Tab. \ref{tab:results}.

\xhdr{How crucial is localization for PointNav?}
We answer this by comparing the upper-bound performance of the agent trained and evaluated with oracle localization (GT-localization) with an agent that has been trained and evaluated to navigate without any form of localization input (the no-localization baseline). We see a drastic drop in both SPL ($0.866$ to $0.096$) and Success ($0.948$ to $0.099$) as the localization input is taken away. This makes it seem like the baseline method is completely failing. However, we observe that these baselines are able to navigate reasonably well, they are just not reaching the exact goal location (SoftSPL of $0.726$ for no-localization vs. $0.865$ for gt-localization). Therefore, to get a meaningful measure of performance, we compare SoftSPL numbers, wherever appropriate in the text that follows.

Next, we also compare the dead-reckoning agent with memorized odometry estimates with the no-localization baseline. We see that the former outperforms the latter (SPL=$0.096$, SoftSPL=$0.726$ vs. SPL=$0.303$, SoftSPL=$0.797$). This shows that having some form of localization (albeit, erroneous)  is essential for navigation. Our proposed agent with learnt odometry prediction) is able to further outperform dead-reckoning with $67.6\%$ relative improvements in SPL ($0.508$ vs. $0.303$).

\xhdr{How privileged is an agent with access to oracle localization at test time?} By comparing the gt-localization agent with our ego-localization approach, we find that our agent succeeds less often (SPL=$0.508$, Success=$0.535$ vs. SPL=$0.866$, Success=$0.948$). However, it is able to make progress towards the goal by a degree that reasonably matches that of the ``oracle'' agent (SoftSPL=$0.813$ v/s SoftSPL=$0.865$). This can be explained by the fact that our agent, with noisy estimates of its location thinks that it has reached the goal and calls stop prematurely (geo\_d\_T=$0.959$ v/s geo\_d\_T=$0.388$).

\xhdr{How well can we do by simply memorizing odometry estimates?}
Recall that, in the absence of any noise in the agent's actuations, memorized egomotion estimates will be incorrect only during collisions. Our agent consistently outperforms dead-reckoning across all navigation metrics in the noiseless settings -- with $2\%$, $67.6\%$, $72\%$ and $8.4\%$ relative improvements in SoftSPL, SPL, Success and geo\_d\_T, respectively. This points to the utility of learning egomotion from visual inputs.

\xhdr{Transferring policies to noisy actuations.}
We shift our attention to the noisy actuation setting now. For this setting, we present results for the following set-ups: (a) old-policy: where we evaluate policies trained in noiseless actuations without any fine-tuning, (b) re-trained policy: where the policies are re-trained and evaluated in the noisy set-up.

As one would expect, when policies trained in a noiseless actuation setting are transferred to environments with noisy actions, navigation performance suffers. For the no-localization, gt-localization and our ego-localization agents, the SoftSPLs drop from $0.726$, $0.865$ and $0.813$ to $0.491$, $0.666$ and $0.280$ (old-policy setting for no-localization, gt-localization and ego-localization in Tab. \ref{tab:results}), respectively. It is interesting to note that this performance drop affects agents with imperfect localization estimates worse than those with oracle localization. This is because, for the former, noisy actuations become all the more challenging -- the agent doesn't end up where it would expect its actions to lead to, and there is no corrective signal that can allow for a potential re-calibration towards the goal. 

Next, we investigate ways to \textit{retrain} our agents to adapt to noise in the actuations. Recall that our proposed agent offers a decoupling between learning dynamics (odometry) and navigation (policy). Taking advantage of this, we treat the policy as a ``swappable'' component that can be used with a different odometry model re-calibrated to noisy actuations. We first fine-tune the odometer on the noisy version of the egomotion dataset (Sec \ref{subsec: noisy-acts}), followed by using the fine-tuned odometer with the frozen policy (the retrained-odom setting in Tab. \ref{tab:results}). We see significant performance gains in doing so -- a relative improvement of $76.43\%$ in SoftSPL ($0.280$, old-policy vs. $0.494$, retrained-odom).

In the absence of a separation between dynamics and policy, the only way to adapt the no-localization and gt-localization agents to the noisy setting is via an expensive re-training of the policy. Doing so leads to performance gains across both sets of agents -- relative improvements of $4.88\%$ and $0.75\%$ in SoftSPL, respectively (old-policy v/s re-trained policy). Moreover, doing the analogous re-training of our agent's policy (with the re-trained odometer) i.e. re-trained-odom+policy, leads us to out-perform all baseline agents in noisy actuation settings as well (SoftSPL=$0.576$ for our proposed agent vs. $0.515$, $0.407$ for no-localization and dead-reckoning, respectively).

\begin{figure}[t]
    \centering
    \includegraphics[scale=0.21, clip, trim=0.6cm 18.5cm 0.5cm 0.5cm]{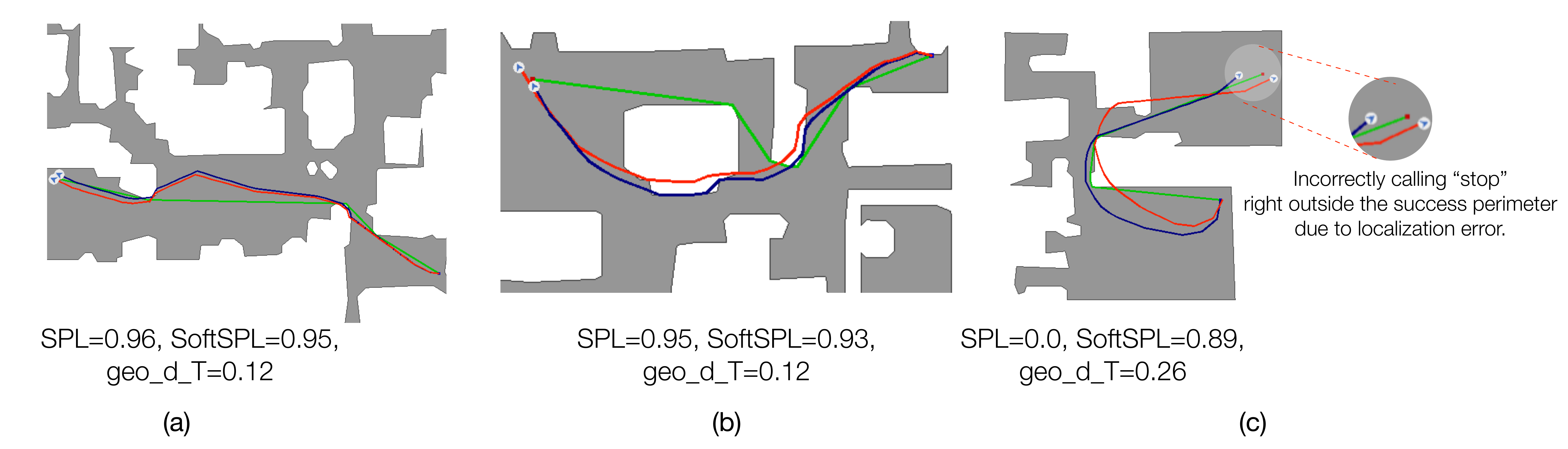}
    \caption{Top-down visualization of our ego-localization agent trajectories. In all cases, the shortest path, is shown in \textcolor{green}{green}. (a), (b) demonstrate successful trajectories where the agent's predicted pose (\textcolor{red}{red}) closely follows its actual trajectory (\textcolor{blue}{blue}). In (c), the agent's pose predictions deviates substantially from its trajectory causing it to incorrectly call \texttt{stop} at the success boundary.}
    \label{fig:top-down-trajs}
\end{figure}

\xhdr{Comparisons with classic approaches.} In the noiseless setting, our learnt visual odometry estimates lead to better PointNav agents than using ORB-SLAM2 \cite{mur2017orb} for localization -- with $39.2\%$, $6.3\%$ and $18.9\%$ relative improvements in SoftSPL, SPL and geo\_d\_T, respectively. With noisy actuations, the baseline has a higher success rate (SPL=$0.267$ vs. $0.047$). However, it takes much longer to reach the goal (SoftSPL=$0.301$ vs. $0.576$). Note that the high success but poor SPL of this baseline is simply a statement about its heuristic path-planning aspect, and not about the performance of SLAM-based localization. Specifically, the baseline uses a set of hand-coded heuristics to select and move towards a waypoint on the planned-path. Every-so-often (with a probability of $0.1$), the agent executes a randomly sampled action (an avenue to recover, in case the agent gets stuck). Therefore, this baseline agent does make progress towards the goal, but in a suboptimal number of steps on account of the above heuristics-based/occasionally-random sampling of actions.


\xhdr{Habitat Challenge 2020}
Our submission, comprising of a simple modification of our proposed agent, to the PointNav track of Habitat Challenge 2020, was ranked \#1 on the Test-Challenge leaderboard with respect to SoftSPL (0.596) and distance to goal (1.824) and \#2 with respect to SPL (0.146). Please check the Supplementary document for details regarding the challenge configuration settings, modifications to our approach and a snapshot of the leaderboard.
\section{Conclusion}
We develop Point-Goal navigation agents that can operate without explicit localization information from the environment (no GPS+Compass sensor) in the presence of noisy transition dynamics. Our agent, with learnt visual odometry modules, demonstrates performance improvements over naive adaptions of existing agents to our setting as well as strong (learnt and traditional) baselines and emerges as runners-up in the Habitat Challenge 2020.  We also show that such a separation between learning the dynamics of the environment (via the odometer) and learning to navigate (via the policy) allows for a straight-forward transfer of our agent from noiseless to noisy actuation settings -- circumventing the time and resource intensive training of near-perfect navigation policies as in prior work \cite{ddppo}. 
We view this as a step towards one of the grand goals of the community -- making navigation agents more suitable for deployment in the real world.

\section{Acknowledgements}
The Georgia Tech effort was supported in part by NSF, AFRL, DARPA, ONR YIPs, ARO PECASE, Amazon and Siemens.
The views and conclusions contained herein are those of the authors and should not be interpreted as necessarily
representing the official policies or endorsements, either expressed or implied, of the U.S. Government, or any sponsor.

\LetLtxMacro{\section}{\oldsection}
\bibliography{bib/main}
\clearpage

\section{Appendix}

\subsection{LoCoBot noise models}
For our experiments with noisy actuations, we source noise models that were collected in the real world by measuring the accuracies of position controllers (implementations of low-level control that the robot uses to get to a desired pose) on a physical robot. The authors in \cite{murali2019pyrobot} experimented with implementations of 3 different controllers on both LoCoBot and LoCoBot-Lite \cite{locobot} -- (1) DWA, (2) Proportional Controller and (3) ILQR (refer to \cite{murali2019pyrobot} for a description of the controllers). Trials in the real world were conducted to quantify the difference in the commanded v/s achieved state of the robot by using each of the above controllers. \cite{murali2019pyrobot} reports that the errors were generally lower for LocoBot and ILQR performed the best among the controllers. Hence, we source the noise models derived from the LoCoBot-ILQR trials for experiments in our paper.

To re-cap, we model both translational and rotational noise in the agent's actuations. Translational noise is measured along $\text{z}$ (the direction of agent motion) and $\text{x}$ (the direction perpendicular to motion) on the ground plane and modelled as a bi-variate Gaussian (with a diagonal co-variance matrix). Rotational noise is measured along rotation around $\text{y}$ and modelled as a uni-variate Gaussian.

Also recall that, for any given action, we add noise to both the agent's location as well as heading (by sampling from the respective noise distributions) in order to simulate say, an agent turning a bit while attempting to move forward (or, vice-versa). We use different sets of translational and rotational noise models for (a) linear motion (i.e. \texttt{move-forward}) and rotational motion (i.e. \texttt{turn-left} and \texttt{turn-right}). We now present the parameters of the noise models. 

\xhdr{Linear motion}. For linear motion (i.e. \texttt{move-forward} action), the mean and co-variance for the translational bi-variate Gaussian noise model are: 
\begin{align*}
\mu = \left[ \mu_z = 0.014, \mu_x = 0.009 \right] \\
\Sigma = \texttt{diag}(0.006, 0.005)
\end{align*}
The mean and variance for the uni-variate rotational noise model are: $\mu = 0.008$ and $\sigma = 0.004$.

\xhdr{Rotational motion}. Similarly, for rotational motion (i.e. \texttt{turn-left}, \\\texttt{turn-right}), the mean and co-variance for the translational bi-variate Gaussian noise model are: 
\begin{align*}
    \mu = \left[ \mu_z = 0.003, \mu_x = 0.003 \right] \\
    \Sigma = \texttt{diag}(0.002, 0.003)
\end{align*}
The mean and variance for the uni-variate rotational noise model are: $\mu = 0.023$ and $\sigma = 0.012$.

\subsection{DD-PPO Training}
Recall that we use DD-PPO \cite{ddppo} to train our agent policies. Following \cite{ddppo}, we force the rollout collection by all DD-PPO workers to end (and model optimization to start) after 80\% (pre-emptive threshold) of the workers have finished collecting rollouts. This is done in order to avoid delays due to ``stragglers’’ and leads to better scaling across multiple GPUs. We use PPO with Generalized Advantage Estimation \cite{schulman2017ppo}. We set the discount factor $\gamma$ to $0.99$, the GAE parameter $\tau$ to $0.95$ and the clip parameter to $0.2$ (along with a linear decay for the clip parameter with the number of PPO updates). We use the Adam optimizer with an initial learning rate of $2.5$e-$4$ and with a linear decay. We clip gradients norms to a max of $0.5$ before policy updates. These hyper-parameters are consistent with the experimental settings in \cite{habitat19iccv}.

\subsection{Hyperparameters for the classic navigation baseline}
The classic navigation baseline (Sec 4 in the main paper) builds a map of size $400 \times 400$ where each grid/cell in the map corresponds to $0.1$m $\times 0.1$m dimensions of physical space. The mapper estimates an occupancy map of the environment from depth maps and camera intrinsics. At any given step, all points in the depth map are projected into an egocentric point cloud, followed by a thresholding operation where only points within a depth range of $0.1$m to $4$m are retained. The point cloud is then transformed to global scene coordinates with the help of pose estimates from the ORB-SLAM2 localization module. This is followed by projecting all points that lie within a range of $\left [0.3, 1 \right]$ times the camera height ($1.25$m) onto the ground plane. To create a 2-D obstacle map, any map cell with at least 128 projected points is treated as an obstacle. 

Given the obstacle map generated by the mapper, the SLAM-estimated pose and the target goal location, the planner finds a shortest path from the agent's location to the goal via the D* Lite algorithm \cite{koenig2002d}. Given that planned path, the baseline selects a waypoint along the path that lies at a distance of at least $0.5$m from the agent's current state.  If the agent's heading is within a range of $15^\circ$ from the direction of this waypoint, it executes a \texttt{move-forward} action. Otherwise, it rotates (\texttt{turn-left} or \texttt{turn-right}) towards the direction of the waypoint. Every so often (with a probability $0.1$), a random action (among \texttt{move-forward}, \texttt{turn-left} and \texttt{turn-right}) is executed.

\begin{figure}[t]
    \centering
        \includegraphics[scale=0.4, clip, trim=2cm 0cm 6cm 0cm]{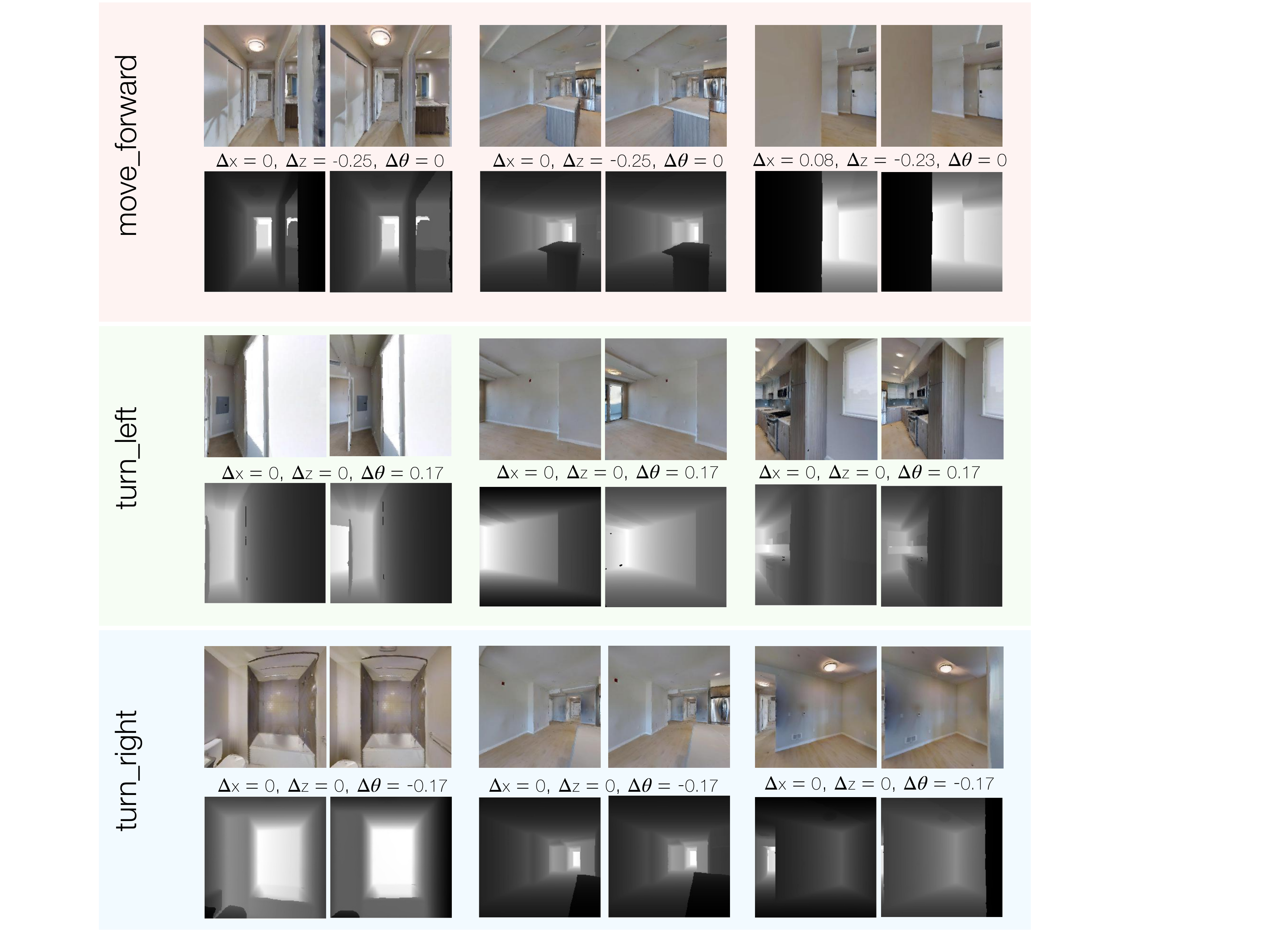}
    \caption{Visualization of samples from the egomotion dataset collected in a \textbf{noiseless} actuation setting. We group the data points according to the action -- \texttt{move-forward}, \texttt{turn-left} and \texttt{turn-right}. For every data point, we show the RGB image frames, depth maps and the relative pose change between the source and target agent states.}
    \label{fig:noiseless-ego-dataset-viz}
\end{figure}

\begin{figure}[t]
    \centering
        \includegraphics[scale=0.4, clip, trim=2cm 0cm 4cm 0cm]{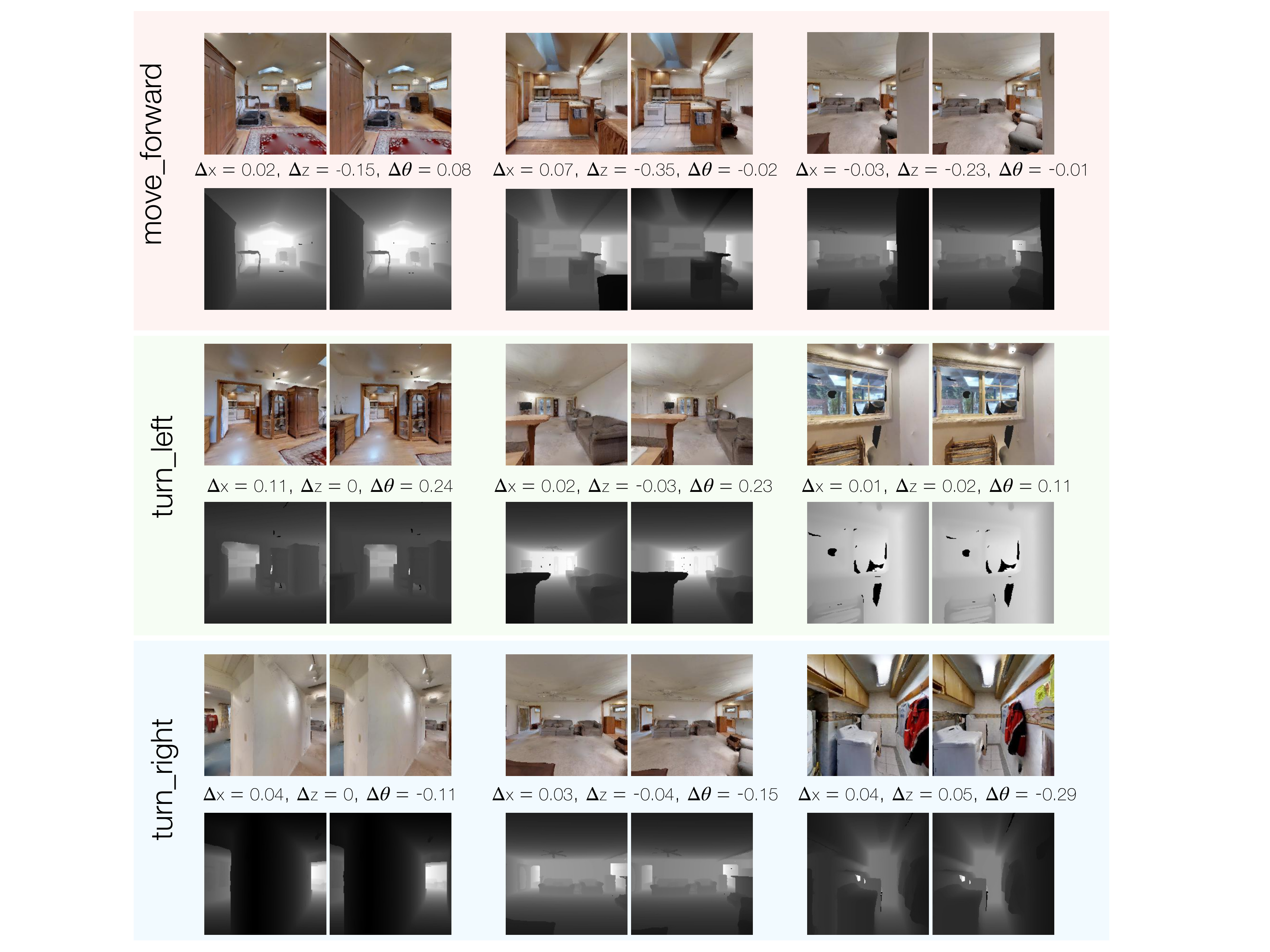}
    \caption{Visualization of samples from the egomotion dataset collected in a \textbf{noisy} actuation setting. We group the data points according to the action -- \texttt{move-forward}, \texttt{turn-left} and \texttt{turn-right}. For every data point, we show the RGB image frames, depth maps and the relative pose change between the source and target agent states.}
    \label{fig:noisy-ego-dataset-viz}
\end{figure}

\subsection{Qualitative examples from the Egomotion dataset}
Fig \ref{fig:noiseless-ego-dataset-viz}. shows some qualitative examples of data points from our egomotion dataset that has been collected in the noiseless actuation setting. For each of the $3$ action classes in our dataset (\texttt{move-forward}, \texttt{turn-left} and \texttt{turn-right}), we show randomly sampled data points. Recall (from Sec. 3 in the main paper) that each data point in our dataset consists of RGB frames and depth maps corresponding to the source (src) and target (tgt) agent states as well as the associated ground-truth relative pose change between src and tgt. Since there's no noise in the agent actuations, the agent always turns by \emph{exactly} $10^\circ$ (or, $0.17$ radians) when executing \texttt{turn-left} or \texttt{turn-right} (and doesn't suffer any displacements in the ground plane while doing so). Similarly, a \texttt{move-forward} doesn't lead to any change in the agent's heading. The right-most example under the \texttt{move-forward} action presents an instance where the agent suffers a collision (with a wall) while attempting to move forward. As a result, the agent also has some non-zero displacement along the x-axis (the direction perpendicular to the agent's heading in the ground plane). Note that the displacement along the z-axis (the agent's heading) is also different from $0.25$m (the default actuation specification for the \texttt{move-forward} action).

\begin{figure}[t]
    \centering
        \includegraphics[scale=0.35, clip, trim=0cm 11cm 0cm 0cm]{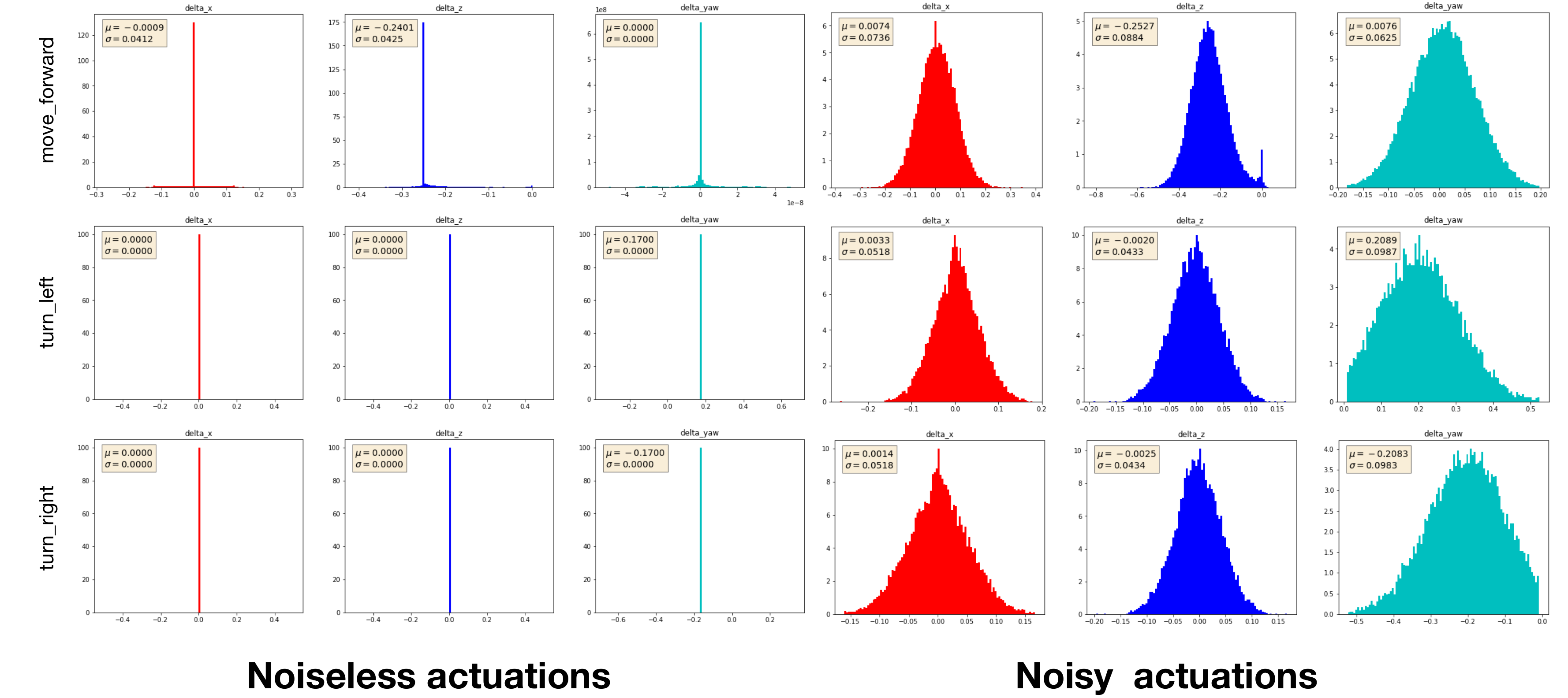}
    \caption{A comparison of the distribution of per-action egomotion deltas between noiseless and noisy actuations settings of our egomotion dataset. For noiseless actuations, turn actions always correspond to a $10^\circ$ turn. Forward actions result in a displacement of $0.25$m along the agent's heading with the exception of collisions where the agent potentially also suffers some displacement along x. In the noisy setting, for every action, we sample and add noise from LoCoBot \cite{locobot} to all 3 degrees of freedom.}
    \label{fig:ego-dataset}
\end{figure}

Fig. \ref{fig:noisy-ego-dataset-viz} presents qualitative examples from the noisy actuation version of the egomotion dataset. Note that irrespective of the type of action -- translational (\texttt{move-forward}) or rotational (\texttt{turn-left} or \texttt{turn-right}), we record non-zero pose changes across x, z as well as $\theta$ (that deviates from the default actuation specifications for the actions). Fig. \ref{fig:ego-dataset} shows a comparison of the per-action egomotion distribution between the noiseless and noisy actuation settings.

\subsection{Standalone evaluation of the visual odometry model}
Recall that, the odometry model that our proposed agent is equipped with is pre-trained on the egomotion dataset and then kept frozen during the training/evaluation of the agent policy. In Sec. 5 of the main paper, we report results for our agent's navigation performance (using our odometer to derive localization estimates). In addition to that, in this section, we also report numbers for a standalone evaluation of our visual odometry model on the task of regressing to the ground-truth relative pose estimates between source and target agent states.

We perform such an evaluation of our odometry model under two settings -- novel observation pairs from previously-seen environments (val-seen) and novel observation pairs from unseen environments (val-unseen). We generate the former by sampling data points from scenes in the training split (and ensuring there is no overlap with the train set of data points) and the latter by generating data points from scenes in the val split. For both splits (val-seen and val-unseen), the data collection protocol remains the same, as described in Sec. 3 in the main paper.

For the odometry model trained on noiseless agent actuations, the Smooth-L1 loss on the val-unseen and val-seen splits are $0.56$e-$4$ and $0.46$e-$4$, respectively. The val-unseen split provides a comparatively more challenging set-up than val-seen due to the added complexity of previously-unseen environments (in addition to evaluation of novel observation pairs) and this is reflected in the trends in the Smooth-L1 loss values.

\begin{figure}[t]
    \centering
        \includegraphics[scale=0.35, clip, trim=0cm 5cm 0cm 0cm]{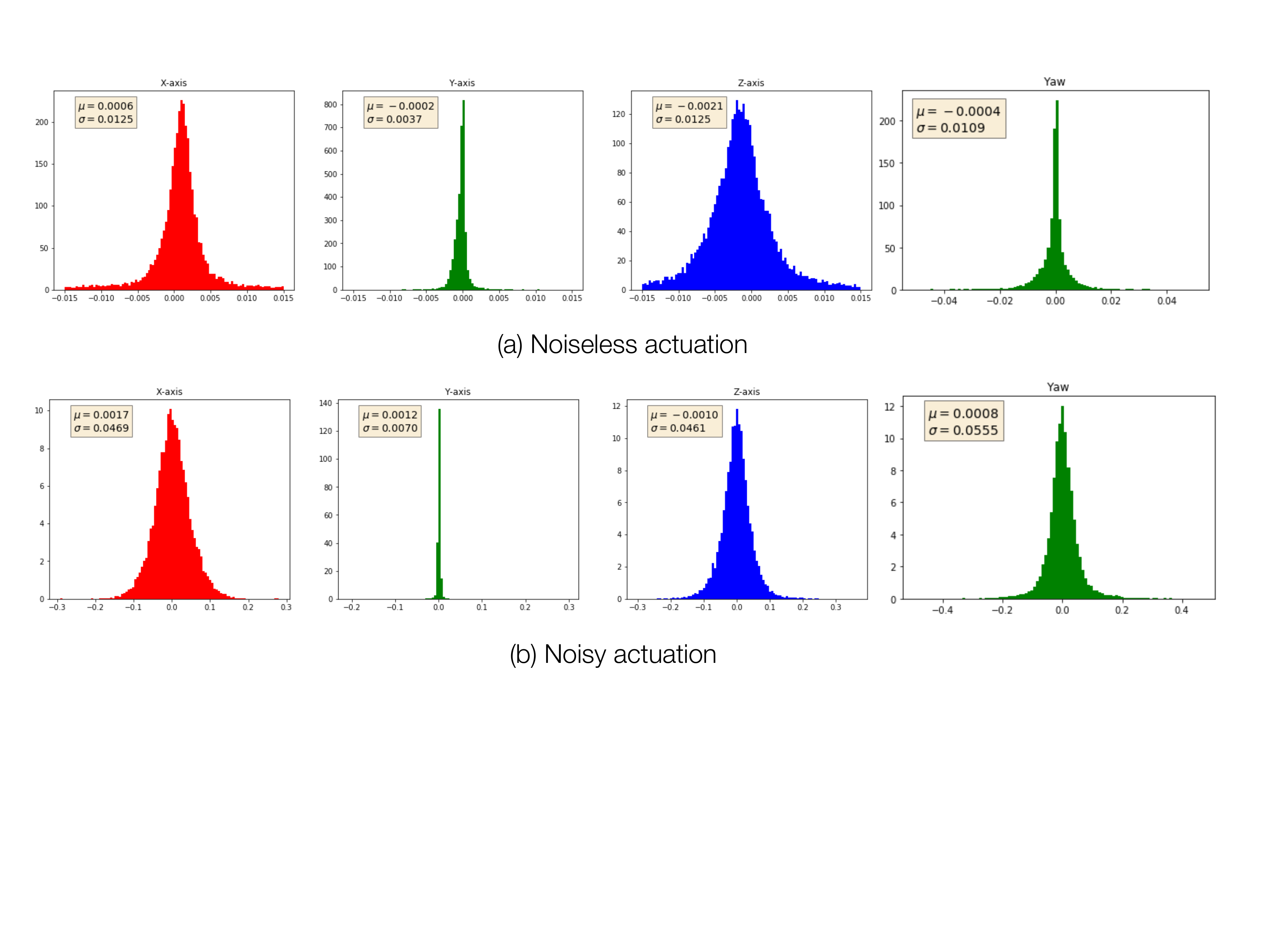}
    \caption{We present the egomotion prediction errors, broken down as errors along x, y, z and yaw for the odometry model trained in (a) noiseless and (b) noisy actuation settings.}
    \label{fig:odom-stats}
\end{figure}

We also report the distribution of egomotion prediction errors factored along x, y, z and yaw ($\theta$) for the val-unseen split in Fig \ref{fig:odom-stats}. Error here refers to the difference in the predicted and the ground-truth values for the x, y, z and yaw components of the relative pose.

Note that although the task of regressing to the relative pose change, given visual observation pairs seems straightforward (as evidenced by the low loss values on val), navigating using these integrated egomotion estimates from the odometry model still remains a challenging task (as seen by the difference in performance between gt-localization and ego-localization in Sec. 5 of the paper). This is due to the fact that although the visual odometry model is able to deliver locally accurate predictions of egomotion, integration of these estimates along trajectories leads to accumulation of errors (Fig. 3 from the main paper).

\subsection{Large-scale training}
The results reported in the paper (Sec. 5) correspond to training our agents (and baselines) for up to 60M frames of experience. We also experiment with large-scale training of our agents for up to 300M frames of experience (across 128 GPUs) in noisy actuation settings. At 300M frames, our proposed agent ego-localization (SoftSPL=0.517, SPL=0.034, geo\_d\_T=2.535) is still able to outperform the no-localization (SoftSPL=0.411, SPL=0.005, geo\_d\_T=3.127) baseline. The ``oracle'' gt-localization navigator's performance at 300M is as follows: (SoftSPL=0.542, SPL=0.594, geo\_d\_T=0.831).

\subsection{Habitat Challenge 2020}
We submitted a straightforward modification of our proposed (ego-localization) agent to the Point-Goal navigation (PointNav) track of Habitat Challenge 2020 (organized as part of the Embodied AI Workshop at CVPR 2020). In addition to noisy agent actuations and the absence of a GPS+Compass sensor, the configuration settings of the challenge also include the following:
\begin{itemize}
    \item visual sensing noise (noisy RGB-D observation maps)
    \item a change in the agent's sliding dynamics. As per the simulator settings used in the experiments reported in the paper (also consistent with prior work \cite{habitat19iccv,wijmans2019embodied}), the agent “slides” along walls and boundaries of obstacles during collisions. This design choice was inspired by game engines where such sliding behavior allows for a smoother player control. However, this behavior is not realistic — a real-world robot would bump into obstacles and simply stop upon colliding.
    \item wider turn angles ($30^\circ$ vs. $10^\circ$)
    \item physical agent dimensions and camera configuration parameters (spatial resolution, field-of-view and camera height) set to match the settings in LoCoBot.
\end{itemize}

\begin{figure}[t]
    \centering
    \includegraphics[scale=0.18, clip, trim=0cm 19cm 0cm 0cm]{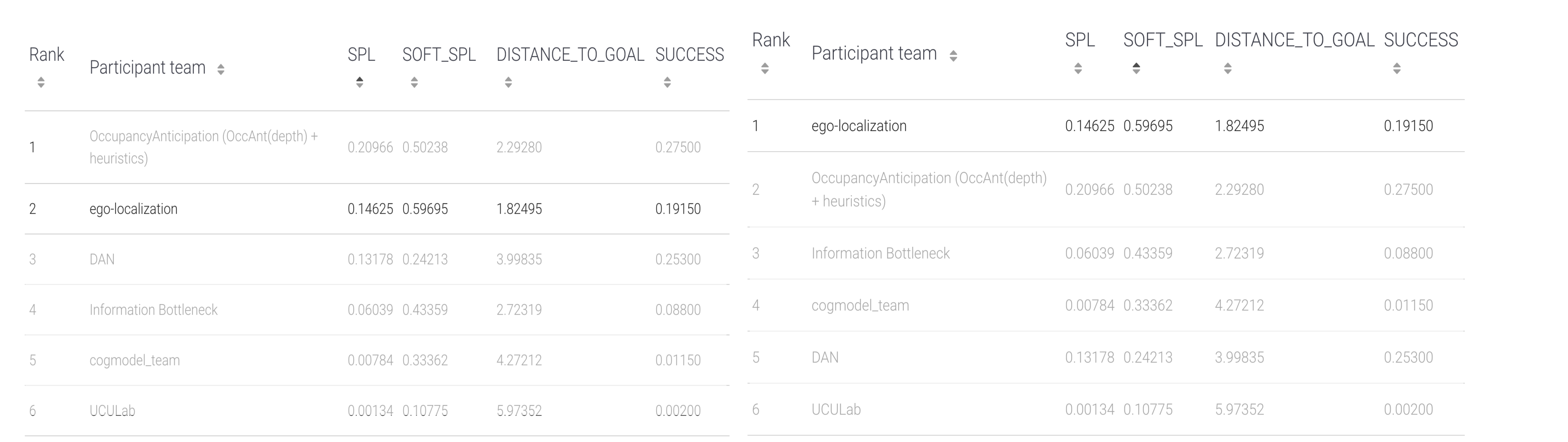}
    \caption{Leaderboard for the challenge phase of the PointNav track in the CVPR 2020 Habitat Challenge. The leaderboard is sorted by SPL on the left and SoftSPL on the right.}
    \label{fig:challenge-leaderboard}
\end{figure}

For the purposes of our challenge submissions, we replaced the 3-layer CNN encoder of our visual odometry model with a ResNet18 based encoder and trained the odometer on a version of the egomotion dataset collected under the challenge settings. We found that using a higher capacity ResNet18 encoder in the odometer was necessary to outperform baselines in the presence of noise in the depth maps. We also re-train our agent policy under the new challenge settings.

The challenge results are shown in Fig. \ref{fig:challenge-leaderboard}. Our submission (under the team name, ``ego-localization'') was ranked \#1 on the Test-Challenge leaderboard with respect to SoftSPL (0.596) and distance to goal (1.824) and \#2 with respect to SPL (0.146).

\end{document}